\documentclass[10pt, a4paper]{article}
\usepackage{lrec}

\usepackage{url}
\usepackage{multirow}
\usepackage{amsfonts}
\usepackage{graphicx} 
\usepackage{amsmath}
\usepackage[ruled,vlined]{algorithm2e}
\usepackage{fancyhdr,amssymb}
\usepackage{enumitem}
\usepackage{xcolor}

\title{Multilingual Twitter Corpus and Baselines for Evaluating Demographic Bias in Hate Speech Recognition}

\name{Xiaolei Huang$^{1*}$, Linzi Xing$^{2}$, Franck Dernoncourt$^{3}$, Michael J. Paul$^{1}$\thanks{~~The work was partially done when the first author worked as an intern at Adobe Research.}}

\address{1.University of Colorado Boulder, 2. University of British Columbia\, 3. Adobe Research\\
1. \{xiaolei.huang, mpaul\}@colorado.edu, 2. lzxing@cs.ubc.ca 3. dernonco@adobe.com
}

\abstract{
Existing research on fairness evaluation of document classification models mainly uses synthetic monolingual data without ground truth for author demographic attributes.
In this work, we assemble and publish a multilingual Twitter corpus for the task of hate speech detection with inferred four author demographic factors: age, country, gender and race/ethnicity. 
The corpus covers five languages: English, Italian, Polish, Portuguese and Spanish.
We evaluate the inferred demographic labels with a crowdsourcing platform, Figure Eight. 
To examine factors that can cause biases, we take an empirical analysis of demographic predictability on the English corpus.
We measure the performance of four popular document classifiers and evaluate the fairness and bias of the baseline classifiers on the author-level demographic attributes.
\\ \newline }

\begin{document}

\maketitleabstract

\section{Introduction}

While document classification models should be objective and independent from human biases in documents, research have shown that the models can learn human biases and therefore be discriminatory towards particular demographic groups~\cite{dixon2018measuring,borkan2019nuanced,sun2019mitigating}.
The goal of \textit{fairness-aware} document classifiers is to train and build non-discriminatory models towards people no matter what their demographic attributes are, such as gender and ethnicity.
Existing research~\cite{dixon2018measuring,kiritchenko2018examining,park2018reducing,garg2019counterfactual,borkan2019nuanced} in evaluating fairness of document classifiers focus on the \textit{group fairness}~\cite{chouldechova2018frontiers}, which refers to every demographic group has equal probability of being assigned to the positive predicted document category.

However, the lack of original author demographic attributes and multilingual corpora bring challenges towards the fairness evaluation of document classifiers.
First, the datasets commonly used to build and evaluate the fairness of document classifiers obtain derived synthetic author demographic attributes instead of the original author information.
The common data sources either derive from Wikipedia toxic comments~\cite{dixon2018measuring,park2018reducing,garg2019counterfactual} or synthetic document templates~\cite{kiritchenko2018examining,park2018reducing}.
The Wikipedia Talk corpus\footnote{\url{https://figshare.com/articles/Wikipedia_Detox_Data/4054689}}~\cite{wulczyn2017ex} provides demographic information of annotators instead of the authors, Equity Evaluation Corpus\footnote{\url{http://saifmohammad.com/WebPages/Biases-SA.html}}~\cite{kiritchenko2018examining} are created by sentence templates and combinations of racial names and gender coreferences.
While existing work~\cite{davidson2019racial,diaz2018addressing} infers user demographic information (white/black, young/old) from the text, such inference is still likely to cause confounding errors that impact and break the independence between demographic factors and the fairness evaluation of text classifiers.
Second, existing research in the fairness evaluation mainly focus on only English resources, such as age biases in blog posts~\cite{diaz2018addressing}, gender biases in Wikipedia comments~\cite{dixon2018measuring} and racial biases in hate speech detection~\cite{davidson2019racial}.
Different languages have shown different patterns of linguistic variations across the demographic attributes~\cite{johannsen2015cross,huang2019neural}, methods~\cite{zhao2017men,park2018reducing} to reduce and evaluate the demographic bias in English corpora may not apply to other languages. 
For example, Spanish has gender-dependent nouns, but this does not exist in English~\cite{sun2019mitigating}; and Portuguese varies across Brazil and Portugal in both word usage and grammar~\cite{maier2014language}.
The rich variations have not been explored under the fairness evaluation due to lack of multilingual corpora.
Additionally, while we have hate speech detection datasets in multiple languages~\cite{waseem2016hateful,sanguinetti2018italian,ptaszynski2017learning,basile2019semeval,fortuna2019hierarchically}, there is still no integrated multilingual corpora that contain author demographic attributes which can be used to measure group fairness.
The lack of author demographic attributes and multilingual datasets limits research for evaluating classifier fairness and developing unbiased classifiers.

In this study, we combine previously published corpora labeled for Twitter hate speech recognition in English~\cite{waseem2016hateful,waseem2016you,founta2018large}, Italian~\cite{sanguinetti2018italian}, Polish~\cite{ptaszynski2017learning}, Portuguese~\cite{fortuna2019hierarchically}, and Spanish~\cite{basile2019semeval}, and publish this multilingual data augmented with author-level demographic information for four attributes: race, gender, age and country.
The demographic factors are inferred from user profiles, which are independent from text documents, the tweets.
To our best knowledge, this is the first \textbf{multilingual} hate speech corpus annotated with \textbf{author attributes} aiming for fairness evaluation.
We start with presenting collection and inference steps of the datasets.
Next, we take an exploratory study on the language variations across demographic groups on the English dataset.
We then experiment with four multiple classification models to establish baseline levels of this corpus.
Finally, we evaluate the fairness performance of those document classifiers.

\section{Data}

We assemble the annotated datasets for hate speech classification.
To narrow down the data sources, we limit our dataset sources to the unique online social media site, Twitter.
We have requested 16 published Twitter hate speech datasets, and finally obtained 7 of them in five languages.
By using the Twitter streaming API\footnote{\url{https://developer.twitter.com/}}, we collected the tweets annotated by hate speech labels and their corresponding user profiles in English~\cite{waseem2016hateful,waseem2016you,founta2018large}, Italian~\cite{sanguinetti2018italian}, Polish~\cite{ptaszynski2017learning}, Portuguese~\cite{fortuna2019hierarchically}, and Spanish~\cite{basile2019semeval}.
We binarize all tweets' labels (indicating whether a tweet has indications of hate speech), allowing to merge the different label sets and reduce the data sparsity.

Whether a tweet is considered hate speech heavily depends on who the speaker is; for example, whether a racial slur is intended as hate speech depends in part on the speaker's race~\cite{waseem2016hateful}.
Therefore, hate speech classifiers may not generalize well across all groups of people, and disparities in the detection offensive speech could lead to bias in content moderation~\cite{ICWSM1817809}.
Our contribution is to further annotate the data with user demographic attributes inferred from their public profiles,
thus creating a corpus suitable for evaluating author-level fairness for this hate speech recognition task across multiple languages.

\subsection{User Attribute Inference}
We consider four user factors of age, race, gender and geographic location. For location, we inference two granularities, country and US region, but only experiment with the country attribute.
While the demographic attributes can be inferred through tweets~\cite{volkova2015inferring,davidson2019racial},
we intentionally exclude the contents from the tweets if they infer these user attributes, in order to make the evaluation of fairness more reliable and independent.
If users were grouped based on attributes inferred from their text, then any differences in text classification across those groups could be related to the same text. 
Instead, we infer attributes from public user profile information (i.e., description, name and photo).

\paragraph{Age, Race, Gender.}
We infer these attributes from each user's profile image by using Face++ (\url{https://www.faceplusplus.com/}),
a computer vision API that provides estimates of demographic characteristics.
Empirical comparisons of facial recognition APIs have found that Face++ is the most accurate tool on Twitter data~\cite{jung2018assessing} and works comparatively better for darker skins~\cite{buolamwini2018gender}.
For the gender, we choose the binary categories (male/female) by the predicted probabilities.
We map the racial outputs into four categories: Asian, Black, Latino and White.
We only keep users that appear to be at least 13 years old, and we save the first result from the API if multiple faces are identified.
We experiment and evaluate with binarization of race and age with roughly balanced distributions (white and nonwhite, $\leq$ median vs. elder age) to consider a simplified setting across different languages, since race is harder to infer accurately.

\paragraph{Country.}
The country-level language variations can bring challenges that are worth to explore.
We extract geolocation information from users whose profiles contained either numerical location coordinates or a well-formatted (matching a regular expression) location name. 
We fed the extracted values to the Google Maps API (\url{https://maps.googleapis.com}) to obtain structured location information (city, state, country).
We first count the main country source and then binarize the country to indicate if a user is in the main country or not. 
For example, the majority of users in the English are from the United States (US), therefore, we can binarize the country attributes to indicate if the users are in the US or not.

\subsection{Corpus Summary} 

We show the corpus statistics in Table~\ref{tab:corpus} and summarize the full demographic distributions in Table~\ref{tab:user}. 
The binary demographic attributes (age, country, gender, race) can bring several benefits. 
First, we can create comparatively balanced label distributions. 
We can observe that there are differences in the race and gender among Italian and Polish data, while other attributes across the other languages show comparably balanced demographic distributions.
Second, we can reduce errors inferred from the Face++ on coarse labels.
Third, it is more convenient for us to analyze, conduct experiments and evaluate the group fairness of document classifiers.

\begin{table}[htp]
\centering
\begin{tabular}{c||cccc}
Language & Users & Docs & Tokens & HS Ratio \\\hline\hline
English & 64,067 & 83,077 & 20.066 & .370 \\
Italian & 3,810 & 5,671 & 19.721 & .195 \\
Polish & 86 & 10,919 & 14.285 & .089 \\
Portuguese & 600 & 1,852 & 18.494 & .205 \\
Spanish & 4,600 & 4,831 & 19.199 & .397
\end{tabular}
\caption{Statistical summary of multilingual corpora across English, Italian, Polish, Portuguese and Spanish. We present number of users (Users), documents (Docs), and average tokens per document (Tokens) in the corpus, 
plus the label distribution (HS Ratio, percent of documents labeled positive for hate speech).}
\label{tab:corpus}
\end{table}

\begin{table*}[htp]
\centering
\begin{tabular}{c||cc|cc|cc|cc}
Language & \multicolumn{2}{c|}{Age} & \multicolumn{2}{c|}{Country} & \multicolumn{2}{c|}{Gender} & \multicolumn{2}{c}{Race} \\\hline\hline
\multirow{2}{*}{English} & Mean & Median & US & non-US & Female & Male & White & non-White \\
 & 32.041 & 29 & .599 & .401 & .499 & .501 & .505 & .495 \\\hline
\multirow{2}{*}{Italian} & Mean & Median & Italy & non-Italy & Female & Male & White & non-White \\
 & 44.518 & 43 & .778 & .222 & .307 & .692 & .981 & .018 \\\hline
\multirow{2}{*}{Polish} & Mean & Median & Poland & non-Poland & Female & Male & White & non-White \\
 & 39.245 & 38 & .795 & .205 & .324 & .676 & .895 & .105 \\\hline
\multirow{2}{*}{Portuguese} & Mean & Median & Brazil & non-Brazil & Female & Male & White & non-White \\
 & 29.635 & 26 & .437 & .563 & .569 & .431 & .508 & .492 \\\hline
\multirow{2}{*}{Spanish} & Mean & Median & Spain & non-Spain & Female & Male & White & non-White \\
 & 31.911 & 27 & .339 & .661 & .463 & .537 & .549 & .451
\end{tabular}
\caption{Statistical summary of user attributes in age, country, gender and race. For the age, we present both mean and median values in case of outliers. For the other attributes, we show binary distributions.}
\label{tab:user}
\end{table*}

Table~\ref{tab:corpus} presents different patterns of the corpus. 
The Polish data has the smallest users. 
This is because the data focuses on the people who own the most popular accounts in the Polish data~\cite{ptaszynski2017learning}, the other data collected tweets randomly.
And the dataset shows a much more sparse distribution of the hate speech label than the other languages.

Table~\ref{tab:user} presents different patterns of the user attributes. 
English, Portuguese and Spanish users are younger than the Italian and Polish users in the collected data.
And both Italian and Polish show more skewed demographic distributions in country, gender and race, while the other datasets show more balanced distributions.

\subsection{Demographic Inference Accuracy}

Image-based approaches will have inaccuracies, as a person's demographic attributes cannot be conclusively determined merely from their appearance. 
However, given the difficulty in obtaining ground truth values, we argue that automatically inferred attributes can still be informative for studying classifier fairness. 
If a classifier performs significantly differently across different groups of users, then this shows that the classifier is biased along certain groupings, even if those groupings are not perfectly aligned with the actual attributes they are named after. 
This subsection tries to quantify how reliably these groupings correspond to the demographic variables.

\begin{table}[th]
\centering
\begin{tabular}{c||c|c|c}
 & Age & Race & Gender \\ \hline
\multicolumn{4}{c}{Annotator Agreement} \\ \hline
Face++ & .80 & .80 & .98 \\ \hline
\multicolumn{4}{c}{Accuracy} \\ \hline
English & .86 & .90 & .94 \\ \hline
Italian & .82 & .96 & .98 \\ \hline
Polish & .88 & .96 & .98 \\ \hline
Portuguese & .82 & .78 & .92 \\ \hline
Spanish & .76 & .82 & .90 \\ \hline
Overall & .828 & .884 & .944 \\ 
\end{tabular}
\caption{Annotator agreement (percentage overlap) and evaluation accuracy for Face++.}
\label{tab:api_eval}
\end{table}

Prior research found that Face++ achieves 93.0\% and 92.0\% accuracy on gender and ethnicity evaluations~\cite{jung2018assessing}.
We further conduct a small evaluation on the hate speech corpus by a small sample of annotated user profile photos providing a rough estimate of accuracy while acknowledging that our annotations are not ground truth.
We obtained the annotations from the crowdsourcing website, Figure Eight (\url{https://figure-eight.com/}).
We randomly sampled 50 users whose attributes came from Face++ in each language.
We anonymize the user profiles and feed the information to the crowdsourcing website.
Three annotators annotated each user photo with the binary demographic categories.
To select qualified annotators and ensure quality of the evaluations, we set up 5 golden standard annotation questions for each language.
The annotators can join the evaluation task only by passing the golden standard questions.
We decide demographic attributes by majority votes and present evaluation results in Table~\ref{tab:api_eval}.
Our final evaluations show that overall the Face++ achieves averaged accuracy scores of 82.8\%, 88.4\% and 94.4\% for age, race and gender respectively.

\subsection{Privacy Considerations}
To facilitate the study of classification fairness, we will publicly distribute this anonymized corpus with the inferred demographic attributes including both original and binarized versions.
To preserve user privacy, we will not publicize the personal profile information, including user ids, photos, geocoordinates as well as other user profile information, which were used to infer the demographic attributes.
We will, however, provide inferred demographic attributes in their original formats from the Face++ and Google Maps based on per request to allow wider researchers and communities to replicate the methodology and probe more depth of fairness in document classification.

\vspace{-.2cm}

\section{Language Variations across\\ Demographic Groups}
Demographic factors can improve the performances of document classifiers~\cite{hovy2015demographic}, and demographic variations root in language, especially in social media data~\cite{volkova2013exploring,hovy2015demographic}.
For example, language styles are highly correlated with authors' demographic attributes, such as age, race, gender and location~\cite{coulmas_2017,preoctiuc2018user}. Research~\cite{bolukbasi2016man,zhao2017men,garg2018word} find that biases and stereotypes exist in word embeddings, which is widely used in document classification tasks. For example, ``receptionist'' is closer to females while ``programmer'' is closer to males, and ``professor'' is closer to Asian Americans while ``housekeeper'' is closer to Hispanic Americans.

This motivates us to explore and test if the language variations hold in our particular dataset, how strong the effects are. 
We conduct the empirical analysis of demographic predictability on the English dataset.

\subsection{Are Demographic Factors Predictable in Documents?}
\label{subsec:pred}

\begin{table*}[htp]
\centering
\begin{tabular}{cc||c}
\multicolumn{2}{c||}{Demographic Attributes} & Top 10 Features of Demographic Attribute Prediction \\\hline\hline
\multirow{2}{*}{Race} & White & nigga, fucking, ass, bro, damn, niggas, sir, moive, melon, bitches \\
 & Other & abuse, gg, feminism, wadhwa, feminists, uh, freebsd, feminist, ve, blocked \\\hline
\multirow{2}{*}{Gender} & Female & rent, driving, tho, adorable, met, presented, yoga, stressed, awareness, me \\
 & Male & idiot, the, players, match, idiots, sir, fucking, nigga, bro, trump
\end{tabular}
\caption{Top 10 predictable features of race and gender in the English dataset.}
\label{tab:features}
\end{table*}

We examine how accurately the documents can predict author demographic attributes from three different levels:
\begin{enumerate}
    \item Word-level. We extract TF-IDF-weighted 1-, 2-grams features.
    \item POS-level. We use Tweebo parser~\cite{kong2014dependency} to tag and extract POS features. We count the POS tag and then normalize the counts for each document.
    \item Topic-level. We train a Latent Dirichlet Allocation~\cite{blei2003latent} model with 20 topics using Gensim~\cite{rehurek2010software} with default parameters. Then a document can be represented as a probabilistic distribution over the 20 topics.
\end{enumerate}

\begin{figure}[htp]
\centering
\includegraphics[width=0.40\textwidth]{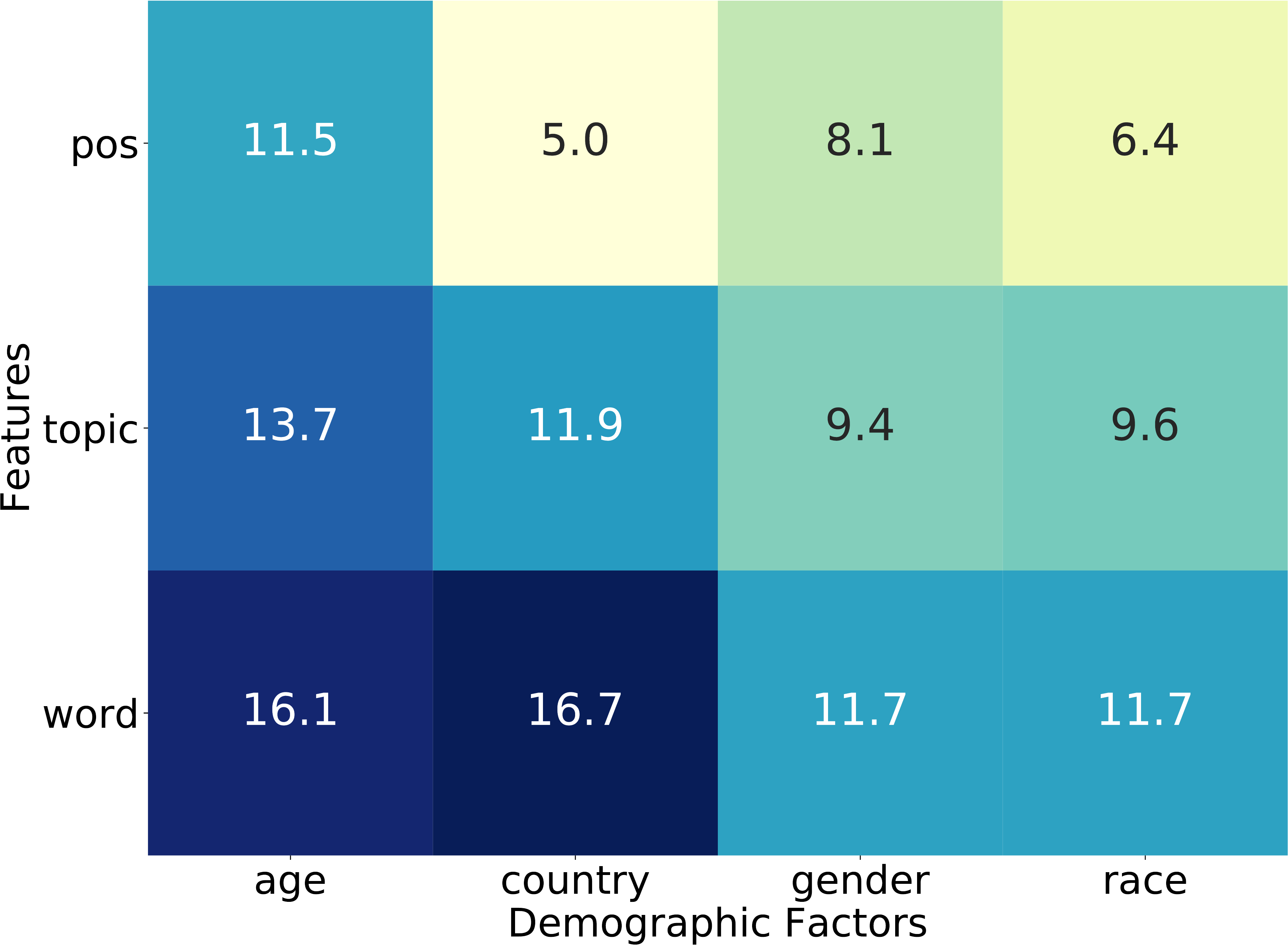}
\caption{Predictability of demographic attributes from the English data. We show the absolute percentage improvements in accuracy over majority-class baselines. The majority-class baselines of accuracy are .500 for the binary predictions. The darker color indicates higher improvements and vice versa.}
\label{fig:predictability}
\end{figure}

We shuffle and split data into training (70\%) and test (30\%) sets.
Three logistic classifiers are trained by the three levels of features separately.
We measure the prediction accuracy and show the absolute improvements in Figure~\ref{fig:predictability}.

The improved prediction accuracy scores over majority baselines suggest that language variations across demographic groups are encoded in the text documents. 
The results show that documents are the most predictable to the age attribute.
We can also observe that the word is the most predictable feature to demographic factors,
while the POS feature is least predictable towards the country factor.
These suggest there might be a connection between language variations and demographic groups.
This motivates us to further explore the language variations based on word features.
We rank the word features by mutual information classification~\cite{pedregosa2011scikit} and present the top 10 unigram features in Table~\ref{tab:features}.
The qualitative results show the most predictable word features towards the demographic groups and 
suggest such variations may impact extracted feature representations and further training fair document classifiers.

The Table~\ref{tab:features} shows that when classifying hate speech tweets, the n-words and b-words are more significant correlated with the white instead of the other racial groups.
However, this shows an opposite view than the existing work~\cite{davidson2019racial}, which presents the two types of words are more significantly correlated with the black.
This can highlight the values of our approach that to avoid confounding errors, we obtain author demographic information independently from the user generated documents.

\section{Experiments}

Demographic variations root in documents, especially in social media data~\cite{volkova2013exploring,hovy2015demographic,johannsen2015cross}.
Such variations could further impact the performance and fairness of document classifiers.
In this study, we experiment four different classification models including logistic regression (LR), recurrent neural network (RNN)~\cite{chung2014empirical}, convolutional neural network (CNN)~\cite{kim2014convolutional} and Google BERT~\cite{devlin2019bert}.
We present the baseline results of both performance and fairness evaluations across the multilingual corpus.

\subsection{Data Preprocessing}
To anonymize user information, we hash user and tweet ids and then replace hyperlinks, usernames, and hashtags with generic symbols (URL, USER, HASHTAG).
Documents are lowercased and tokenized using NLTK~\cite{bird2004nltk}. 
The corpus is randomly split into training (70\%), development (15\%), and test (15\%) sets. 
We train the models on the training set and find the optimal hyperparameters on the development set before final evaluations on the test set. 
We randomly shuffle the training data at the beginning of each training epoch.

\begin{table*}[htp]
\centering
\begin{tabular}{cc|cccc}
Language & Method & Acc & F1-w & F1-m & AUC \\\hline
\multirow{4}{*}{English} & LR & .874 & .874 & .841 & .920 \\
 & CNN & .878 & .877 & .845 & .927 \\
 & RNN & \textbf{.898} & \textbf{.896} & \textbf{.867} & \textbf{.938} \\
 & BERT & .705 & .635 & .579 & .581
\end{tabular}
\quad
\begin{tabular}{cc|cccc}
Language & Method & Acc & F1-w & F1-m & AUC \\\hline
\multirow{4}{*}{Italian} & LR & .660 & .679 & .631 & .725 \\
 & CNN & .687 & .702 & .651 & .745 \\
 & RNN & \textbf{.729} & \textbf{.731} & \textbf{.666} & \textbf{.763} \\
 & BERT & .697 & .629 & .468 & .498
\end{tabular}

\begin{tabular}{cc|cccc}
\multicolumn{6}{c}{} \\
Language & Method & Acc & F1-w & F1-m & AUC \\\hline
\multirow{4}{*}{Polish} & LR & \textbf{.864} & .846 & .653 & .804 \\
 & CNN & .855 & .851 & .688 & .813 \\
 & RNN & .857 & \textbf{.854} & \textbf{.696} & \textbf{.822} \\
 & BERT & .824 & .782 & .478 & .474
\end{tabular}
\quad
\begin{tabular}{cc|cccc}
\multicolumn{6}{c}{} \\
Language & Method & Acc & F1-w & F1-m & AUC \\\hline
\multirow{4}{*}{Portuguese} & LR & .660 & .598 & .551 & .648 \\
 & CNN & \textbf{.681} & \textbf{.674} & \textbf{.653} & \textbf{.719} \\
 & RNN & .607 & .586 & .553 & .633 \\
 & BERT & .613 & .568 & .525 & .524
\end{tabular}

\begin{tabular}{cc|cccc}
\multicolumn{6}{c}{} \\
Language & Method & Acc & F1-w & F1-m & AUC \\\hline
\multirow{4}{*}{Spanish} & LR & \textbf{.704} & \textbf{.707} & \textbf{.698} & \textbf{.761} \\
 & CNN & .650 & .654 & .645 & .710\\
 & RNN & .674 & .674 & .658 & .720 \\
 & BERT & .605 & .573 & .502 & .505
\end{tabular}
\caption{Overall performance evaluation of baseline classifiers. We evaluate overall performance by four metrics including accuracy (Acc), weighted F1 score (F1-w), macro F1 score (F1-m) and area under the ROC curve (AUC). The higher score indicates better performance. We highlight models achieve the best performance in each column.}
\label{tab:perform}
\end{table*}

\subsection{Baseline Models}
We implement and experiment four baseline classification models. 
To compare fairly, we keep the feature size up to 15K for each classifier across all five languages.
We calculate the weight for each document category by $\frac{N}{N_l}$~\cite{king2001logistic}, where $N$ is the number of documents in each language and $N_l$ is the number of documents labeled by the category.
Particularly, for training BERT model, we append two additional tokens, ``[CLS]'' and ``[SEP]'', at the start and end of each document respectively.
For the neural models, we pad each document or drop rest of words up to 40 tokens.
We use ``unknown'' as a replacement for unknown tokens.
We initialize CNN and RNN classifiers by pre-trained word embeddings~\cite{mikolov2013distributed,godin2015multimedia,bojanowski2017enriching,deriu2017leveraging} and train the networks up to 10 epochs.

\paragraph{LR.} 
We first extract TF-IDF-weighted features of uni-, bi-, and tri-grams on the corpora, using the most frequent 15K features with the minimum feature frequency as 2. 
We then train a \texttt{LogisticRegression} from scikit-learn~\cite{pedregosa2011scikit}. 
We use ``liblinear'' as the solver function and leave the other parameters as default.

\paragraph{CNN.} 
We implement the Convolutional Neural Network (CNN) classifier described in~\cite{kim2014convolutional,zimmerman2018improving} by Keras~\cite{chollet2015keras}.
We first apply 100 filters with three different kernel sizes, 3, 4 and 5.
After the convolution operations, we feed the concatenated features to a fully connected layer and output document representations with 100 dimensions.
We apply ``softplus'' function with a l2 regularization with $.03$ and a dropout rate with $.3$ in the dense layer.
The model feeds the document representation to final prediction.
We train the model with batch size 64, set model optimizer as Adam~\cite{kingma2014adam} and calculate loss values by the cross entropy function.
We keep all other parameter settings as described in the paper~\cite{kim2014convolutional}.

\paragraph{RNN.}
We build a recurrent neural network (RNN) classifier by using bi-directional Gated Recurrent Unit (bi-GRU)~\cite{chung2014empirical,park2018reducing}.
We set the output dimension of GRU as 200 and apply a dropout on the output with rate $.2$.
We optimize the RNN with RMSprop~\cite{tieleman2012lecture} and use the same loss function and batch size as the CNN model.
We leave the other parameters as default in the Keras~\cite{chollet2015keras}.

\paragraph{BERT}
BERT is a transformer-based pre-trained language model which was well trained on multi-billion sentences publicly available on the web~\cite{devlin2019bert}, which can effectively generate the precise text semantics and useful signals.
We implement a BERT-based classification model by HuggingFace's Transformers~\cite{Wolf2019HuggingFacesTS}.
The model encodes each document into a fixed size (768) of representation and feed to a linear prediction layer.
The model is optimized by \texttt{AdamW} with a warmup and learning rate as $.1$ and $2e^{-5}$ respectively.
We leave parameters as their default, conduct fine-tuning steps with 4 epochs and set batch size as 32~\cite{sun2019fine}.
The classification model loads ``bert-base-uncased'' pre-trained BERT model for English and ``bert-base-multilingual-uncased'' multilingual BERT model~\cite{gertner2019mitre} for the other languages.
The multilingual BERT model follows the same method of BERT by using Wikipedia text from the top 104 languages.
Due to the label imbalance shown in Table~\ref{tab:corpus}, we balance training instances by randomly oversampling the minority during the training process.

\subsection{Evaluation Metrics}

\paragraph{Performance Evaluation.}
To measure overall performance, we evaluate models by four metrics: accuracy (Acc), weighted F1 score (F1-w), macro F1 score (F1-m) and area under the ROC curve (AUC). %
The F1 score coherently combines both precision and recall by $2*\frac{precision*recall}{precision+recall}$.
We report F1-m considering that the datasets are imbalanced.

\paragraph{Fairness Evaluation.}
To evaluate {group fairness}, we measure the \textit{equality differences} (ED) of true positive/negative and false positive/negative rates for each demographic factor. 
ED is a standard metric to evaluate fairness and bias of document classifiers~\cite{dixon2018measuring,park2018reducing,garg2019counterfactual}.

This metric sums the differences between the rates within specific user groups and the overall rates.
Taking the false positive rate (FPR) as an example, we calculate the equality difference by:
$$FPED = \sum_{d \in D}|FPR_d - FPR|$$
, where $D$ is a demographic factor (e.g., race) and $d$ is a demographic group (e.g., white or nonwhite).

\section{Results}
We have presented our evaluation results of performance and fairness in Table~\ref{tab:perform} and Table~\ref{tab:fairness} respectively.
Country and race have very skewed distributions in the Italian and Polish corpora, therefore, we omit fairness evaluation on the two factors.

\begin{table*}[htp]
\centering
\begin{tabular}{cc|ccc}
\multicolumn{5}{c}{\textbf{Age}} \\\hline\hline
Language & Method & FNED & FPED & SUM-ED\\\hline
\multirow{4}{*}{English} & LR & .059 & .104 & .163\\
 & CNN  & .052 & .083 & .135 \\
 & RNN  & .041 & .118 & .159 \\
 & BERT & .004 & .012 & .016 
\end{tabular}
\quad
\begin{tabular}{cc|ccc}
\multicolumn{5}{c}{\textbf{Gender}} \\\hline\hline
Language & Method & FNED & FPED & SUM-ED \\\hline
\multirow{4}{*}{English} & LR & .023 & .056 & .079 \\
 & CNN  & .018 & .056 & .074 \\
 & RNN  & .013 & .055 & .068 \\
 & BERT & .007 & .009 & .016
\end{tabular}

\begin{tabular}{cc|ccc}
\multicolumn{5}{c}{} \\\hline\hline
Language & Method & FNED & FPED & SUM-ED \\\hline
\multirow{4}{*}{Italian} & LR & .076 & .194 & .270\\
 & CNN  & .003 & .211 & .214\\
 & RNN  & .042 & .185 & .227\\
 & BERT & .029 & .034 & .063
\end{tabular}
\quad
\begin{tabular}{cc|ccc}
\multicolumn{5}{c}{} \\\hline\hline
Language & Method & FNED & FPED & SUM-ED \\\hline
\multirow{4}{*}{Italian} & LR  & .145 & .020 & .165 \\
 & CNN  & .064 & .094 & .158 \\
 & RNN  & .088 & .075 & .163 \\
 & BERT & .041 & .056 & .097
\end{tabular}

\begin{tabular}{cc|ccc}
\multicolumn{5}{c}{} \\\hline\hline
Language & Method & FNED & FPED & SUM-ED \\\hline
\multirow{4}{*}{Polish} & LR & .256 & .059 & .315 \\
 & CNN  & .389 & .138 & .527 \\
 & RNN  & .335 & .089 & .424 \\
 & BERT & .027 & .027 & .054
\end{tabular}
\quad
\begin{tabular}{cc|ccc}
\multicolumn{5}{c}{} \\\hline\hline
Language & Method & FNED & FPED & SUM-ED \\\hline
\multirow{4}{*}{Polish} & LR & .266 & .045 & .309 \\
 & CNN  & .411 & .048 & .459 \\
 & RNN  & .340 & .034 & .374 \\
 & BERT & .042 & .013 & .055
\end{tabular}

\begin{tabular}{cc|ccc}
\multicolumn{5}{c}{} \\\hline\hline
Language & Method & FNED & FPED & SUM-ED \\\hline
\multirow{4}{*}{Portuguese} & LR  & .061 & .044 & .105 \\
 & CNN  & .033 & .096 & .129 \\
 & RNN  & .079 & .045 & .124 \\
 & BERT & .090 & .097 & .187 
\end{tabular}
\quad
\begin{tabular}{cc|ccc}
\multicolumn{5}{c}{} \\\hline\hline
Language & Method & FNED & FPED & SUM-ED \\\hline
\multirow{4}{*}{Portuguese} & LR  & .052 & .007 & .059 \\
 & CNN  & .018 & .013 & .031 \\
 & RNN  & .099 & .083 & .182 \\
 & BERT & .055 & .125 & .180
\end{tabular}

\begin{tabular}{cc|ccc}
\multicolumn{5}{c}{} \\\hline\hline
Language & Method & FNED & FPED & SUM-ED \\\hline
\multirow{4}{*}{Spanish} & LR & .089 & .013 & .102 \\
 & CNN  & .117 & .139 & .256 \\
 & RNN  & .078 & .083 & .161 \\
 & BERT & .052 & .015 & .067
\end{tabular}
\quad
\begin{tabular}{cc|ccc}
\multicolumn{5}{c}{} \\\hline\hline
Language & Method & FNED & FPED & SUM-ED \\\hline
\multirow{4}{*}{Spanish} & LR & .131 & .061 & .292 \\
 & CNN  & .032 & .108 & .140 \\
 & RNN  & .030 & .039 & .069 \\
 & BERT & .021 & .016 & .037
\end{tabular}

\begin{tabular}{cc|ccc}
\multicolumn{5}{c}{} \\
\multicolumn{5}{c}{} \\
\multicolumn{5}{c}{\textbf{Country}} \\\hline\hline
Language & Method & FNED & FPED & SUM-ED \\\hline
\multirow{4}{*}{English} & LR & .026 & .053 & .079 \\
 & CNN  & .027 & .063 & .090\\
 & RNN  & .024 & .061 & .085\\
 & BERT & .006 & .001 & .007
\end{tabular}
\quad
\begin{tabular}{cc|ccc}
\multicolumn{5}{c}{} \\
\multicolumn{5}{c}{} \\
\multicolumn{5}{c}{\textbf{Race}} \\\hline\hline
Language & Method & FNED & FPED & SUM-ED \\\hline
\multirow{4}{*}{English} & LR  & .019 & .056 & .075 \\
 & CNN  & .007 & .029 & .036 \\
 & RNN  & .008 & .063 & .071 \\
 & BERT & .003 & .009 & .012 
\end{tabular}

\begin{tabular}{cc|ccc}
\multicolumn{5}{c}{} \\\hline\hline
Language & Method & FNED & FPED \\\hline
\multirow{4}{*}{Portuguese} & LR & .093 & .026 & .119\\
 & CNN  & .110 & .122 & .232 \\
 & RNN  & .022 & .004 & .026 \\
 & BERT & .073 & .071 & .144
\end{tabular}
\quad
\begin{tabular}{cc|ccc}
\multicolumn{5}{c}{} \\\hline\hline
Language & Method & FNED & FPED & SUM-ED \\\hline
\multirow{4}{*}{Portuguese} & LR  & .068 & .005 & .073 \\
 & CNN  & .056 & .033 & .089 \\
 & RNN  & .074 & .054 & .128 \\
 & BERT & .045 & .186 & .231
\end{tabular}

\begin{tabular}{cc|ccc}
\multicolumn{5}{c}{} \\\hline\hline
Language & Method & FNED & FPED & SUM-ED \\\hline
\multirow{4}{*}{Spanish} & LR & .152 & .154 & .306\\
 & CNN  & .089 & .089 & .178 \\
 & RNN  & .071 & .113 & .184 \\
 & BERT & .017 & .017 & .034
\end{tabular}
\quad
\begin{tabular}{cc|ccc}
\multicolumn{5}{c}{} \\\hline\hline
Language & Method & FNED & FPED & SUM-ED \\\hline
\multirow{4}{*}{Spanish} & LR & .095 & .030 & .125 \\
 & CNN  & .072 & .054 & .126 \\
 & RNN  & .011 & .004 & .015 \\
 & BERT & .046 & .005 & .051
\end{tabular}
\caption{Fairness evaluation of baseline classifiers across the five languages on the four demographic factors. We measure fairness and bias of document classifiers by equality differences of false negative rate (FNED), false positive rate (FPED) and sum of FNED and FPED (SUM-ED). The higher score indicates lower fairness and higher bias and vice versa.}
\label{tab:fairness}
\end{table*}

\paragraph{Overall performance evaluation.}
Table~\ref{tab:perform} demonstrates the performances of the baseline classifiers for hate speech classification on the corpus we proposed. 
Results are obtained from the five languages covered in our corpus respectively.
Among the four baseline classifiers, LR, CNN and RNN consistently perform well on all languages.
Moreover, neural-based models (CNN and RNN) substantially outperform LR on four out of five languages (except Spanish).
However, the results obtained by BERT are relatively lower than the other baselines, and show more significant gap in the English dataset.
One possible explanation is BERT was pre-trained on Wikipedia documents, which are significantly different from the Twitter corpus in document length, word usage and grammars.
For example, each tweet is a short document with 20 tokens, but the BERT is trained on long documents up to 512 tokens.
Existing research suggests that fine-tuning on the multilingual corpus can further improve performance of BERT models~\cite{sun2019fine}.

\paragraph{Group fairness evaluation.}

We have measured the group fairness in Table~\ref{tab:fairness}. 
Generally, the RNN classifier achieves better and more stable performance across major fairness evaluation tasks.
By comparing the different baseline classifiers, we can find out that the LR usually show stronger biases than the neural classification models among majority of the tasks.
While the BERT classifier performs comparatively lower accuracy and F1 scores, the classifier has less biases on the most of the datasets.
However, biases can significantly increases for the Portuguese dataset when the BERT classifier achieves better performance.
We examine the relationship by building linear model between two differences: the performance differences between the RNN and other classifiers, the SUM-ED differences between RNN and other classifiers.
We find that the classification performance does not have significantly ($p-value > .05$) correlation with fairness and bias.
The significant biases of classifiers varies across tasks and languages: the classifiers trained on Polish and Italian are biased the most by Age and Gender, the classifiers trained on Spanish and Portuguese are most biased the most by Country, and the classifiers trained on English tweets are the most unbiased throughout all the attributes.
Classifiers usually have very high bias scores on both gender and age in Italian and Polish data.
We find that the age and gender both have very skewed distributions in the Italian and Polish datasets. 
Overall, our baselines provide a promising start for evaluating future new methods of reducing demographic biases for document classification under the multilingual setting.

\section{Conclusion}
In this paper, we propose a new multilingual dataset covering four author demographic annotations (age, gender, race and country) for the hate speech detection task.
We show the experimental results of several popular classification models in both overall and fairness performance evaluations. 
Our empirical exploration indicates that language variations across demographic groups can lead to biased classifiers.
This dataset can be used for measuring fairness of document classifiers along author-level attributes and exploring bias factors across multilingual settings and multiple user factors.
The proposed framework for inferring the author demographic attributes can be used to generate more large-scale datasets or even applied to other social media sites (e.g., Amazon and Yelp).
While we encode the demographic attributes into categories in this work, 
we will provide inferred probabilities of the demographic attributes from Face++ to allow for broader research exploration.
Our code, anonymized data and data statement~\cite{bender2018data} will be publicly available at \url{https://github.com/xiaoleihuang/Multilingual_Fairness_LREC}.

\subsection{Limitations}
While our dataset provides new information on author demographic attributes, and our analysis suggest directions toward reducing bias, a number of limitations must be acknowledged in order to appropriately interpret our findings.

First, inferring user demographic attributes by profile information can be risky due to the accuracy of the inference toolkit.
In this work, we present multiple strategies to reduce the errors bringing by the inference toolkits, such as human evaluation, manually screening and using external public profile information (Instagram).
However, we cannot guarantee perfect accuracy of the demographic attributes,
and, errors in the attributes may themselves be ``unfair'' or unevenly distributed due to bias in the inference tools~\cite{buolamwini2018gender}.
Still, obtaining individual-level attributes is an important step toward understanding classifier fairness, and our results found biases across these groupings of users, even if some of the groupings contained errors.

Second, because methods for inferring demographic attributes are not accurate enough to provide fine-grained information, our attribute categories are still too coarse-grained (binary age groups and gender, and only four race categories).
Using coarse-grained attributes would hide the identities of specific demographic groups, including other racial minorities and people with non-binary gender.
Broadening our analyses and evaluations to include more attribute values may require better methods of user attribute inference or different sources of data.

Third, language variations across demographic groups might introduce annotation biases. Existing research~\cite{sap2019risk} shows that annotators are more likely to annotate tweets containing African American English words as hate speech.
Additionally, the nationality and educational level might also impact on the quality of annotations~\cite{founta2018large}.
Similarly, different annotation sources of our dataset (which merged two different corpora) might have variations in annotating schema.
To reduce annotation biases due to the different annotating schema,
we merge the annotations into the two most compatible document categories: normal and hate speech.
Annotation biases might still exist, therefore, 
we will release our original anonymized multilingual dataset for research communities.

\section{Acknowledgement}
The authors thank the anonymous reviews for their insightful comments and suggestions.
This work was supported in part by the National Science Foundation under award number IIS-1657338.
This work was also supported in part by a research gift from Adobe.

\section{Bibliographical References}
\label{main:ref}
\bibliographystyle{lrec}
\bibliography{reference}

\end{document}